%% file: main.tex
\documentclass{article}

    \usepackage[final,nonatbib]{neurips_2025}

\usepackage[utf8]{inputenc} 
\usepackage[T1]{fontenc}    
\usepackage{hyperref}       
\usepackage{url}            
\usepackage{booktabs}       
\usepackage{amsfonts}       
\usepackage{nicefrac}       
\usepackage{microtype}      
\usepackage{xcolor}         

\usepackage{algorithm}
\usepackage{algorithmic}
\usepackage{csquotes}
\usepackage{amsmath}
\usepackage{enumitem}
\usepackage{booktabs}
\usepackage{tabularx}
\usepackage{adjustbox}
\usepackage{multirow}
\usepackage{pgfplots}
\usepgfplotslibrary{groupplots}
\pgfplotsset{compat=1.18}
\usepackage{hyperref}
\usepackage{xcolor}
\usepackage{listings}
\usepackage{ulem}

\usepackage{subcaption} 

\newcommand{\kc}[1]{{\color{black}#1}}

\usepackage{amssymb}
\usepackage{mathtools}
\usepackage{amsthm}
\theoremstyle{plain}
\newtheorem{theorem}{Theorem}[section]

\theoremstyle{definition}

\theoremstyle{remark}

\title{GVPO: Group Variance Policy Optimization for \\Large Language Model Post-Training}

\author{%
  Kaichen Zhang$^{1,2}$ \quad Yuzhong Hong$^{2}$ \quad Junwei Bao$^{2,}$\thanks{Corresponding Authors. Code available at https://github.com/jszkc/GVPO} \quad Hongfei Jiang$^{2}$ \\
  \textbf{Yang Song$^{2}$} \quad \textbf{Dingqian Hong$^{2}$} \quad \textbf{Hui Xiong$^{1,3,*}$} \\
  $^{1}$Thrust of Artificial Intelligence, Hong Kong University of Science and Technology (Guangzhou) \\
  $^{2}$Zuoyebang Education Technology \quad  $^{3}$Department of Computer Science and Engineering, HKUST \\
  \texttt{kzhangbi@connect.ust.hk} \quad \texttt{eugene.h.git@gmail.com} \quad \texttt{baojunwei001@gmail.com} \\
  \texttt{$\{$jianghongfei,songyang,hongdingqian$\}$@zuoyebang.com} \quad \texttt{xionghui@ust.hk} 
}

\begin{document}
\maketitle

\input{sections/0.abstract}
\input{sections/1.introduction}

\input{sections/3.preliminary}
\input{sections/4.main}
\input{sections/6.experiment}
\input{sections/2.related}
\input{sections/7.conclusions}

\newpage

\begin{ack}
We thank all the reviewers, the AC and program committee members for constructive feedback.

This work was supported in part by the National Key R\&D Program of China (Grant No.2023YFF0725001), in part by the National Natural Science Foundation of China (Grant No.92370204), in part by the guangdong Basic and Applied Basic Research Foundation (Grant No.2023B1515120057), in part by the Education Bureau of Guangzhou.
\end{ack}

\bibliography{main}

\newpage
\input{sections/90.appendix}
\end{document}

%% file: sections/0.abstract.tex
\vspace{-5mm}
\begin{abstract}
Post-training plays a crucial role in refining and aligning large language models to meet specific tasks and human preferences. 
While recent advancements in post-training techniques, such as Group Relative Policy Optimization (GRPO), leverage increased sampling with relative reward scoring to achieve superior performance, these methods often suffer from training instability that limits their practical adoption.
\kc{As a next step}, we present \textbf{Group Variance Policy Optimization (GVPO)}. 
GVPO incorporates the analytical solution to KL-constrained reward maximization directly into its gradient weights, ensuring alignment with the optimal policy. The method provides intuitive physical interpretations: its gradient mirrors the mean squared error between the central distance of implicit rewards and that of actual rewards. GVPO offers two key advantages: (1) it guarantees a \textbf{unique optimal solution}, exactly the KL-constrained reward maximization objective, (2) it supports flexible sampling distributions that \textbf{avoids importance sampling and on-policy limitations}. By unifying theoretical guarantees with practical adaptability, GVPO establishes a new paradigm for reliable and versatile LLM post-training.
\end{abstract}

%% file: sections/1.introduction.tex
\section{Introduction}
Large language models (LLMs) \cite{zhao2025surveylargelanguagemodels,minaee2025largelanguagemodelssurvey}, trained on extensive datasets, exhibit impressive general-purpose capabilities, yet their practical utility and alignment with human values depend critically on post-training \cite{tie2025surveyposttraininglargelanguage} refinement. While pre-training \cite{zhou2023comprehensivesurveypretrainedfoundation, qian2025beyond} equips LLMs with broad linguistic patterns, post-training techniques—such as supervised fine-tuning (SFT) \cite{ouyang2022training} and reinforcement learning \cite{miao2024optimizing} from human feedback (RLHF) \cite{bai2022training} are indispensable for adapting these models to specialized applications and ensuring their outputs align with ethical, safety, and user-centric standards.

\begin{displayquote}
    ``The biggest lesson that can be read from 70 years of AI research is that general methods that leverage computation are ultimately the most effective, and by a large margin.'' \quad\quad\quad\quad\quad\quad\quad\quad\quad\quad\quad\quad--- Rich Sutton, 2024 Turing Award winner
\end{displayquote}

This principle outlined in \textit{The Bitter Lesson} \cite{sutton2019bitter}—which advocates for scalable, computation-driven approaches—is exemplified by recent advances in post-training, particularly Group Relative Policy Optimization (GRPO) \cite{shao2024deepseekmath}. Diverging from conventional reinforcement learning frameworks \cite{schulman2017proximal} that depend on training a separate value function, GRPO directly optimizes advantage by standardizing reward scores across samples. This approach eliminates the need for an auxiliary value model, which typically demands computational resources comparable to those of the policy model itself. As a result, GRPO significantly reduces memory and computational overhead, enabling more efficient sampling and scalable training. 
\kc{Deepseek-R1 \cite{guo2025deepseek} leverages GRPO and achieves significant performance.}

\begin{figure}[t]
\centering
\includegraphics[width=\textwidth]{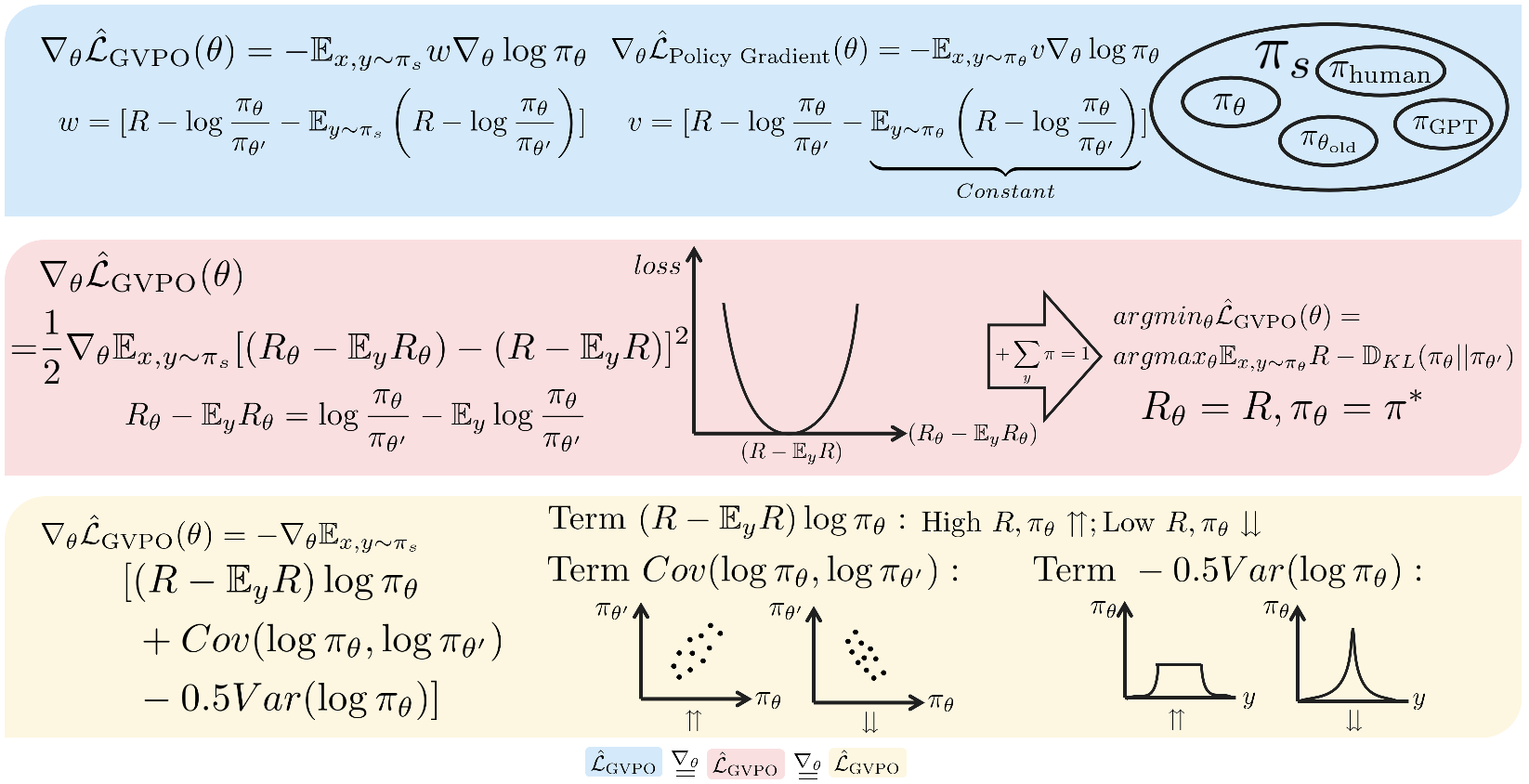}
\vspace*{-5mm}
\caption{\textbf{Three equivalent loss functions of GVPO offer distinct interpretations}: (1) The Negative log-Likelihood perspective (top) illustrates that GVPO accommodates broader sampling distributions compared to conventional policy gradient methods; (2) The Mean Squared Error interpretation (middle) reveals GVPO’s unique optimal solution, which simultaneously maximizes reward under a KL constraint; and (3) The Reinforcement Learning viewpoint (bottom) highlights GVPO’s implicit regularization terms that ensure stable policy optimization. We assume $\beta=1$ for simplicity.}
\label{fig:three}
\vspace{-5mm}
\end{figure}

However, GRPO has been documented to experience issues with training instability in prior literature \cite{yu2025dapo, liu2025understanding}. Specifically, GRPO is highly sensitive to its hyperparameters, such as the clip threshold and the KL-divergence coefficient. 
These limitations undermine the robustness of GRPO and hinder its broader practical adoption.
\kc{As a next step}, we propose \textbf{Group Variance Policy Optimization (GVPO)}, a novel approach for reliable and versatile LLM post-training.

Our analysis begins with a key observation: post-training algorithms—including but not limited to SFT, Reject Sampling \cite{touvron2023llama}, and GRPO—share a unified mathematical structure in their loss gradients \cite{shao2024deepseekmath, gao2024towards}. Specifically, each method’s gradient can be expressed as a weighted sum of the gradients of the log-likelihoods of responses. 
This unified framework reveals that we can directly design weights to encode preferences—positive weights amplify gradients for favored responses, while negative weights suppress disfavored ones, with magnitudes modulating the strength of preference.

Motivated by the success of Direct Preference Optimization (DPO) \cite{rafailov2023direct}—which utilizes a closed-form link between reward models and the optimal policy under KL-divergence constraints \cite{jaques2017sequence}—we explore how to leverage this analytical relationship. A central obstacle arises from the partition function in the closed-form formula, which requires intractable expectation calculations over all possible responses. To address this, we identify a critical condition: 
\textit{when the sum of assigned response weights within a prompt group equals zero, the partition function becomes invariant across compared responses},
effectively canceling out in the policy update rule. This insight eliminates the need for explicit estimation of the partition function, thereby enabling deployment of the closed-form optimal policy while retaining its theoretical advantages.

Based on the previous findings, we design GVPO's weighting scheme where the gradient weight of a response in a group is the difference between the central distance of implicit rewards-which derive from the current policy and the reference policy-and that of actual rewards, illustrated in Figure~\ref{fig:three} (top panel). The loss is computable because the sum of weights in a prompt group equals zero.

We demonstrate that GVPO loss function carries physically meaningful interpretations. Specifically, we establish that its gradient equals that of a mean squared error loss measuring the discrepancy between implicit and actual reward central distances, illustrated in Figure~\ref{fig:three} (middle panel).

Furthermore, the loss function in GVPO can be decomposed into three distinct components, as visualized in Figure~\ref{fig:three} (bottom panel): (1) a group-relative reward term, (2) the variance of the current policy, and (3) the covariance between the current policy and a reference policy. The first component directly promotes advantage maximization by prioritizing responses with higher expected returns. The covariance term acts as a regularizer, mitigating excessive deviations from the reference policy to ensure stable policy updates. Meanwhile, the variance term encourages moderate entropy, thereby naturally balancing exploration and exploitation. We systematically analyze GVPO's structural similarities with conventional policy gradient reinforcement learning methods.

We demonstrate that GVPO offers two key advantages:
\begin{itemize}[leftmargin=0.5cm]
\item GVPO has a unique optimal solution, which coincides precisely with the optimal solution of the KL-constrained reward maximization. This guarantee confers a significant theoretical advantage over DPO. Prior work \cite{bong2022generalized,hong2024energybasedpreferencemodeloffers} highlights that DPO may fail to converge to the optimal policy for the KL-constrained reward maximization problem, because of the inherent flaw of Bradley-Terry model \cite{wu2022asymptoticcomparisonidentifyingconstraints}.
 In contrast, GVPO guarantees that its loss function is aligned with the original constrained optimization problem, ensuring convergence to the globally optimal policy. This theoretical robustness positions GVPO as a more reliable method for policy optimization.
 
\item GVPO supports flexible sampling distributions that avoids importance sampling and on-policy limitations. Beyond the common practice of sampling from the previous step's policy, GVPO retains theoretical guarantees for the unique optimal solution under any sampling distribution satisfying a mild condition.
This property provides a notable theoretical advantage over policy gradient methods \cite{sutton1999policy}. Unlike on-policy approaches \cite{singh2000convergence,williams1992simple}, which require fresh trajectories for updates, GVPO facilitates off-policy training using reusable or heterogeneous datasets. Furthermore, in contrast to off-policy methods \cite{shao2024deepseekmath,schulman2017proximal} reliant on importance sampling, GVPO inherently avoids gradient explosion risks without introducing bias through clipping techniques. 
\end{itemize}

As a result, GVPO emerges as a competitive online RL algorithm, capable of leveraging diverse data sources, sustaining stable policy updates, and preserving convergence to optimality.

%% file: sections/3.preliminary.tex
\section{Preliminary}
Large language models take a prompt $x$ as input and generate a response $y$ as output. A policy $\pi_{\theta}(y_t|x,y_{<t})$ with parameter $\theta$ maps a sequence of tokens generated ($x$ and $y_{<t}$) to a probability distribution over the next token $y_t$. We also denote $\pi_{\theta}(y|x)$ as the probability of generating the response $y$ from $x$. A reward model $R(x,y)$ scores the response $y$ as the reply to the prompt $x$.

A reward model can explicitly be evaluation ratings of human beings; or a trainable function that implicitly reflects human preferences; or a predefined rule, such as correctness, accuracy.

The general purpose of post-training of large language model is summarized as following: Given an initial policy $\pi_{\theta_{init}}$, a dataset of prompts $x \sim D$, a reward model $R$, 
the objective is to train a new policy $\pi_{\theta}$ that generates responses with higher rewards, that is, maximize $E_{x\sim D,y\sim \pi_{\theta}(\cdot|x)}R(x,y)$.

\subsection{Towards better computation leverage in post-training}

The initial stage of large language model post-training typically involves Supervised Fine-Tuning (SFT) \cite{ouyang2022training}. In this phase, a dataset comprising input prompts $x$ paired with exemplary responses $y$ is used to optimize the pre-trained model. The training minimizes the negative log-likelihood loss:
\begin{equation}
\mathcal{L}_{\text{SFT}}(\theta) = -\sum_{(x, y) \in \mathcal{D}} \log \pi_\theta(y|x)
\end{equation}
\kc{
Recent advancements, such as GRPO \cite{shao2024deepseekmath, zhang2025rspo}, better leverage the computation by incorporating multiple sampled responses with their standardized reward scores as weights:
\begin{equation}
\mathcal{L}_{GRPO}(\theta) = -\sum_{(x,\{y_i\}) \in \mathcal{D}} \sum_{i=1}^k Clip(\frac{\pi_\theta(y_i|x)}{\pi_\text{old}(y_i|x)})\frac{R_i-\overline{R}}{\sigma(R)}\log \pi_\theta(y_i|x)- \beta KL(\pi_\theta||\pi_\text{ref})
\end{equation}
Rewards below average lead to negative weights, minimizing their likelihoods. However, minimizing likelihood can result in unstable training \cite{abdolmaleki2024learning}. Additionally, off-policy training of GRPO, which involves importance sampling with the weight $\frac{\pi_\theta}{\pi_{\theta_{old}}}$, becomes unstable when $\pi_\theta$ significantly deviates from $\pi_{\theta_{\text{ref}}}$, potentially causing gradient explosions. To mitigate these issues, GRPO uses gradient clipping and a KL constraint between the updated and reference policies. Nevertheless, empirical results \cite{yu2025dapo,hu2025reinforce++} show that GRPO still exhibits training instability, which undermines its performance. 
}

\subsection{Optimal solution to the KL-constrained reward maximization}
In human preference alignment scenario \cite{ouyang2022training}, the ideal reward model would directly reflect human evaluative judgments. However, obtaining explicit human ratings is often unavailable in practice. Instead, contemporary approaches typically leverage pairwise response preferences $(x,y_w,y_l)$, where $y_w$ denotes the preferred response and $y_l$ the dispreferred response to prompt $x$, to approximate human preferences through reward model training. 
The resulting reward model subsequently enables policy optimization through the following KL-regularized objective:
\begin{equation}
\label{equation:rewardkl}
max_{\pi_{\theta}} \mathbb{E}_{x\sim\mathcal{D},y\sim\pi_\theta(y|x)}[R(x,y)]-\beta\mathbb{D}_{KL}[\pi_\theta(y|x)||\pi_{\theta^\prime}(y|x)]
\end{equation}
where $\beta > 0$ controls the divergence penalty from policy $\pi_{\theta^\prime}$. In preference alignment scenario, $\pi_{\theta^\prime}$ is set to a reference policy $\pi_{ref}$.

Rather than employing separate reward modeling and policy optimization stages, DPO \cite{rafailov2023direct} derives a single-stage training paradigm by exploiting the analytical relationship between optimal policies and reward functions. The optimal solution to Equation~\ref{equation:rewardkl} satisfies:
\begin{equation}
\label{equation:optimal}
\pi^*(y|x)=\frac{1}{Z(x)}\pi_{\theta^\prime}(y|x)e^{R(x,y)/\beta}
\end{equation}
which implies the corresponding reward function:
\begin{equation}
\label{equation:Rxy}
R(x,y)=\beta \log\frac{\pi^*(y|x)}{\pi_{\theta^\prime}(y|x)}+\beta \log Z(x)
\end{equation}
where $Z(x)=\sum_y\pi_{\theta^\prime}(y|x)e^{R(x,y)/\beta}$ represents the partition function. DPO circumvents explicit computation of $Z(x)$ by substituting the reward expression from Equation~\ref{equation:Rxy} into the Bradley-Terry loss \cite{bradley1952rank}, yielding the final objective:
\begin{equation}
\label{equation:DPO}
\mathcal{L}_{\text{DPO}}(\theta) = -\sum_{(x, y_w, y_l) \in \mathcal{D}}\log \sigma(\beta \log\frac{\pi_\theta(y_w|x)}{\pi_{ref}(y_w|x)}- \beta \log\frac{\pi_\theta(y_l|x)}{\pi_{ref}(y_l|x)})
\end{equation}
The success of DPO has been proven to be both efficient and effective. We attribute its achievements to its direct incorporation of the optimal policy's closed-form solution into the training objective.

\subsection{Unified framework of post-training}
As far as we know, post-training algorithms share a unified framework \cite{shao2024deepseekmath, gao2024towards}, in which their losses' gradients share a same format:
\begin{equation}
\nabla_\theta\mathcal{L}(\theta) = -\sum_{(x, y_1,y_2,..,y_k) \in \mathcal{D}} \sum_{i=1}^k w_i \nabla_\theta\log \pi_\theta(y_i|x) \label{equation:pt_framework}
\end{equation}
SFT only has one response per prompt, and its $w_1=1$. 
GRPO's \kc{essential} weights are the standard scores of its rewards in a prompt group. Though it is not obvious for DPO, its gradients also share the same format, in which $w_w = \sigma(\beta \log\frac{\pi_\theta(y_l|x)}{\pi_{ref}(y_l|x)}- \beta \log\frac{\pi_\theta(y_w|x)}{\pi_{ref}(y_w|x)})$ and $w_l=-w_w$. Such the unified framework of post-training holds, because of the chain rule of derivatives.

\input{sections/tables/alg_gvpo}

%% file: sections/tables/alg_gvpo.tex
\begin{algorithm}
\caption{Group Variance Policy Optimization}
\label{alg:gvpo}
\begin{algorithmic}[1]
\REQUIRE initial policy $\pi_{\theta}$; prompt distribution $\mathcal{D}$; hyperparameter $\beta$
    \FOR{step $= 1, \dots, n$}
        \STATE Sample a batch $\mathcal{D}_b$ from $\mathcal{D}$
        \STATE Update the old policy model $\pi_{\theta_{old}} \gets \pi_{\theta}$
        \STATE Sample $k$ responses $\{y_i\}_{i=1}^{k} \sim \pi_{s}(\cdot|x)$ for each prompt $x \in \mathcal{D}_b$
        \STATE Compute rewards $\{R(x,y_i)\}_{i=1}^{k}$ for every sampled response $y_i$ and prompt $x$
        \kc{\STATE Iteratively update policy $\pi_{\theta}$ by minimizing the GVPO loss (Equation~\ref{equation:sample_gvpo}, setting $\pi_{\theta^\prime}=\pi_{\theta_{old}}$)}
    \ENDFOR
\STATE \textbf{Return} $\pi_{\theta}$
\end{algorithmic}
\end{algorithm}

%% file: sections/4.main.tex
\newpage
\section{Group Variance Policy Optimization}
\subsection{Motivation} 
\kc{

The unified post-training framework (Equation~\ref{equation:pt_framework}) indicates that response preferences can be directly incorporated through the assignment of weights $w_i$.

To determine appropriate weights, we draw inspiration from the success of DPO. In particular, we aim to exploit the closed-form relationship between rewards and the optimal solution to the KL-constrained reward maximization objective: $R_\theta(x,y)=\beta \log(\pi_\theta(y|x)/\pi_{\theta^\prime}(y|x))+\beta \log Z(x)$. 

However, the closed-form formula contains a partition function $Z(x)$ that is expensive to estimate in practice, because the function requires calculating the expectation of all possible responses.

To address this issue, we identify a critical condition: when the sum of assigned response weights within a prompt group equals zero, $\sum_{i=1}^k w_i=0$, the partition function becomes invariant across responses: $\sum_{i=1}^k w_iR_\theta(x,y_i)= \sum_{i=1}^k w_i\beta\log(\pi_\theta(y_i|x)/\pi_{\theta^\prime}(y_i|x))$.

}

\subsection{Method} \label{section:method}
Build on this insight, we propose Group Variance Policy Optimization (GVPO), whose gradient weight $w_i$ is the difference between the central distance of implicit rewards-which derive from policy $\pi_{\theta}$ and policy $\pi_{\theta^\prime}$-and that of actual rewards. Formally, GVPO's gradient $\nabla_\theta\mathcal{L}_{\text{GVPO}}(\theta) =$
\begin{equation}
\label{equation:sample_gvpo}
-\beta\sum_{(x, \{y_i\}) \in \mathcal{D}} \sum_{i=1}^k [(R(x,y_i)-\overline{R(x,\{y_i\})})-\beta(\log\frac{\pi_\theta(y_i|x)}{\pi_{\theta^\prime}(y_i|x)} -\overline{\log\frac{\pi_\theta(\{y_i\}|x)}{\pi_{\theta^\prime}(\{y_i\}|x)}})] \nabla_\theta\log \pi_\theta(y_i|x)
\end{equation}
where $\overline{R(x,\{y_i\})}=\frac{1}{k}\sum_{i=1}^k R(x,y_i)$, and $\overline{\log\frac{\pi_\theta(\{y_i\}|x)}{\pi_{\theta^\prime}(\{y_i\}|x)}}=\frac{1}{k}\sum_{i=1}^k\log\frac{\pi_\theta(y_i|x)}{\pi_{\theta^\prime}(y_i|x)}$. We note that GVPO's gradient satisfies $\sum_{i=1}^k w_i=0$. Algorithm~\ref{alg:gvpo} shows our proposed algorithm.

We demonstrate that GVPO's objective carries physically meaningful interpretations: 
\begin{equation}
\begin{aligned}
\nonumber
&\nabla_\theta\mathcal{L}_{\text{GVPO}}(\theta) \\
=&-\sum_{x, \{y_i\} } \sum_{i=1}^k [(R(x,y_i)-\overline{R(x,\{y_i\})})-(R_\theta(x,y_i)-\overline{R_\theta(x,\{y_i\})})] \nabla_\theta \beta\log \pi_\theta(y_i|x) \\
=&-\sum_{x, \{y_i\}} \sum_{i=1}^k [(R(x,y_i)-\overline{R(x,\{y_i\})})-(R_\theta(x,y_i)-\overline{R_\theta(x,\{y_i\})})] \nabla_\theta R_\theta(x,y_i) \\
=&-\sum_{x, \{y_i\} } \sum_{i=1}^k [(R(x,y_i)-\overline{R(x,\{y_i\})})-(R_\theta(x,y_i)-\overline{R_\theta(x,\{y_i\})})] \nabla_\theta(R_\theta(x,y_i)-\overline{R_\theta(x,\{y_i\})}) \\
=& \frac{1}{2}\nabla_\theta\sum_{x, \{y_i\} } \sum_{i=1}^k [(R_\theta(x,y_i)-\overline{R_\theta(x,\{y_i\})})-(R(x,y_i)-\overline{R(x,\{y_i\})})]^2 \\
\end{aligned}
\end{equation}
\kc{The first and second steps hold because $\beta\log Z(x)$ can cancel out. }
The second step holds because $\sum_{i=1}^k w_i \nabla_\theta \overline{R(x,\{y_i\})} =0$. The third step holds because $\nabla_xf(x)^2=2f(x)\nabla_xf(x)$.

Essentially, we have established that GVPO's gradient mathematically equals that of a mean squared error loss measuring the discrepancy between implicit and actual reward central distances. Intuitively, when implicit rewards equal actual rewards or with a constant group shift, the GVPO's loss is minimized. This interpretation also implies that the response with higher actual rewards in a group is also encouraged to have higher implicit rewards, indicating higher $\log\frac{\pi_\theta(y_i|x)}{\pi_{\theta^\prime}(y_i|x)}$.

\kc{Furthermore, by rearranging the mean squared error loss, we can derive a variance-based formulation, which represents the "\textbf{Variance}" term in the name GVPO:}
\[
 \frac{1}{2}\nabla_\theta\sum_{x, \{y_i\} } \sum_{i=1}^k [(R_\theta(x,y_i)-R(x,y_i))-\overline{R_\theta(x,\{y_i\})-R(x,\{y_i\})}]^2
\]

\input{sections/4.3.theory}

\input{sections/4.4.DPO}

\input{sections/4.5.GRPO}
\input{sections/4.6.TRPO}
\input{}

%% file: sections/4.3.theory.tex
\subsection{Theoretical guarantee}

We show that GVPO has an unique optimal solution, and this unique optimal solution is exactly the optimal solution of reward maximization with KL constraint (Equation \ref{equation:optimal}). Formally,
\begin{theorem}
\label{theorem:gvpo}
The unique optimal policy that minimizes $\hat{\mathcal{L}}_{\text{GVPO}}(\theta)$, defined as
\begin{equation}
\label{equation:gvpo}
\hat{\mathcal{L}}_{\text{GVPO}}(\theta)=
\mathbb{E}_{x \sim \mathcal{D}} \mathbb{E}_{y \sim \pi_{s}(\cdot|x)}[(R_\theta(x,y)-
\mathbb{E}_{y \sim \pi_{s}}R_\theta(x,y))-(R(x,y)-
\mathbb{E}_{y \sim \pi_{s}}R(x,y))]^2
\end{equation}
, is given by $\pi_\theta (y|x)=\pi^* (y|x)= \frac{1}{Z(x)}\pi_{\theta^\prime}(y|x)e^{R(x,y)/\beta}$ 
for $\pi_s=\pi_{\theta^\prime}$.
\end{theorem}

\kc{We prove the theorem by establishing both necessity and sufficiency, provided in Appendix~\ref{app:proof1}.}

Theorem~\ref{theorem:gvpo} implies that the parameters minimizing $\hat{\mathcal{L}}_{\text{GVPO}}(\theta)$—guaranteed to be the sole global optimum—also maximize the expected rewards while maintaining proximity to a reference policy. 

The uniqueness of the solution ensures the optimization landscape is well-behaved, avoiding suboptimal local minima and guaranteeing convergence to a single, interpretable policy that optimally balances reward maximization with behavioral consistency relative to the reference. Consequently, this theorem bridges GVPO's practical algorithmic performance with theoretical guarantees.

\begin{theorem}
\label{corollary:pi_s}
The Theorem~\ref{theorem:gvpo} also holds for any sampling distribution $\pi_s$ satisfying $\forall x,\{y|\pi_{\theta^\prime}(y|x)>0\}\subseteq\{y|\pi_s(y|x)>0\}.$
\end{theorem}
\kc{
Beyond the conventional practice of sampling from the reference policy ($\pi_s=\pi_{\theta^\prime}$), GVPO retains the theoretical guarantee of a unique optimal solution under any sampling distribution that satisfies a mild condition. This condition is readily met by any policy $\pi$ where $\pi(y,x)>0$, a criterion inherently fulfilled by contemporary LLM policies utilizing softmax decoding.

The Theorem~\ref{corollary:pi_s} of GVPO opens a new methodological avenue for off-policy LLM post-training. Prior off-policy methods have relied heavily on importance sampling \cite{tokdar2010importance}, which suffers from two key limitations:
(1) when $\pi_\theta$ diverges substantially from $\pi_s$, the importance weight $\frac{\pi_\theta}{\pi_s}$ can become either excessively large or vanishingly small, destabilizing training; and
(2) when sampling involves heuristic or non-parametric components, $\pi_s$ becomes intractable to compute, thereby prohibiting techniques such as experience replay \cite{fedus2020revisiting} in modern LLM post-training. In contrast, GVPO supports highly flexible off-policy sampling strategies while maintaining strong theoretical guarantees, enabling more robust and practical post-training paradigms for large language models.

}

%% file: sections/4.4.DPO.tex
\subsection{Discussions with DPO}
We begin by analyzing the foundational commonality between GVPO and DPO: both methods integrate the closed-form solution to the reward maximization problem under a KL divergence constraint into their training objectives. This integration establishes a direct relationship between the learned policy $\pi_\theta$ and the implicit reward function $R_\theta$, yielding two key advantages:
\begin{itemize}[leftmargin=0.5cm]
    \item It ensures an optimization process that inherently respects the KL divergence constraint, thereby preventing excessive deviation of the policy $\pi_\theta$ from the reference policy $\pi_{ref}$.
    \item It reduces the joint optimization over policies and rewards to a simpler problem focused solely on rewards. The latter is more tractable, as it requires only aligning the implicit rewards $R_\theta(x, y)$ with the true reward function $R(x, y)$.
\end{itemize}

To design effective methods leveraging this closed-form solution, two critical insights emerge:

\begin{enumerate} [leftmargin=0.5cm]
    \item \textbf{Computational Tractability}: The method must avoid intractable terms such as the partition function $Z(x)$. For instance, a naive loss $\mathcal{L} = \sum (R_\theta(x, y) - R(x, y))^2$ fails because $R_\theta(x, y)$ implicitly depends on $Z(x)$, which is computationally infeasible to estimate. DPO circumvents this by adopting the Bradley-Terry preference model, where $Z(x)$ cancels out in pairwise comparisons. GVPO proposes a novel \textit{zero-sum property} across groups of responses, enabling cancellation of $Z(x)$ in broader multi-sample scenarios.
    
    \item \textbf{Alignment with Desired Optimality}: The loss function must enforce meaningful convergence. For example, minimizing $\mathcal{L} = \sum \left(\beta \log \frac{\pi_\theta(x, y)}{\pi_{ref}(x, y)} - R(x, y)\right)^2$ yields a suboptimal solution $R_\theta(x, y) = R(x, y) + \beta \log Z(x)$, which deviates from the true reward $R(x, y)$. A well-designed objective must avoid such misalignment. The method should adapt to available supervision. DPO leverages pairwise preference data without explicit rewards, while GVPO generalizes to group-wise responses with reward signals.
\end{enumerate}

\kc{Moreover, GVPO demonstrates stronger theoretical robustness compared to DPO:}
\begin{itemize}[leftmargin=0.5cm]
    \item Prior work \cite{bong2022generalized,hong2024energybasedpreferencemodeloffers} highlights that DPO may fail to converge to the optimal policy for the KL-constrained reward maximization problem, because of the inherent flaw of Bradley-Terry model \cite{wu2022asymptoticcomparisonidentifyingconstraints}. This arises because the DPO loss admits multiple minimizers, and its correlation with the true reward objective can diminish during training \cite{tang2024generalized}.
    \item In contrast, as formalized in Theorem~\ref{theorem:gvpo} and Theorem~\ref{corollary:pi_s}, GVPO guarantees that its loss function is aligned with the original constrained optimization problem, ensuring convergence to the globally optimal policy. This theoretical robustness positions GVPO as a more reliable method for policy optimization in practice.
\end{itemize}

%% file: sections/4.5.GRPO.tex
\subsection{Discussions with GRPO and Policy Gradient methods} \label{sec:reg}
\kc{\textbf{Structural similarities}}. Seeing the forest for the tree, we compare GVPO not only with GRPO but also with the broader family of policy gradient-based RL methods, beginning with their \kc{superficial} structural similarities. 

For simplicity, we assume \(\beta = 1\) without loss of generality. Then $\hat{\mathcal{L}}_{\text{GVPO}}(\theta)\overset{\nabla_\theta}{=}$
\begin{equation}
\label{equation:gvpovsgrpo}
\begin{aligned}
&\mathbb{E}_{x \sim \mathcal{D},y \sim \pi_{s}(\cdot|x)}
[(R_\theta(x,y) - \mathbb{E}_y R_\theta(x,y))^2-2(R(x,y) - \mathbb{E}_y R(x,y))R_\theta(x,y)] \\
\overset{\nabla_\theta}{=}& \mathbb{E}_{x,y}
[Var(\log\pi_\theta)-2Cov(\log\pi_\theta,\log\pi_{\theta^\prime})-2(R(x,y) - \mathbb{E}_y R(x,y))\log\pi_\theta(y|x)] \\
=&-2\mathbb{E}_{x,y}
[(R(x,y) - \mathbb{E}_y R(x,y))\log\pi_\theta(y|x)+Cov(\log\pi_\theta,\log\pi_{\theta^\prime})-
0.5Var(\log\pi_\theta)]
\end{aligned}
\end{equation}
where $Var(\log\pi_\theta)=(\log\pi_\theta(y|x)-\mathbb{E}_y\log\pi_\theta(y|x))^2$ and $Cov(\log\pi_\theta,\log\pi_{\theta^\prime})=(\log\pi_\theta(y|x)-\mathbb{E}_y\log\pi_\theta(y|x))(\log\pi_{\theta^\prime}(y|x)-\mathbb{E}_y\log\pi_{\theta^\prime}(y|x))$. As shown in Equation~\ref{equation:gvpovsgrpo},
\begin{itemize}[leftmargin=0.5cm]
\item the term \((R(x,y) - \mathbb{E}_y R(x,y))\log\pi_\theta(y|x)\) encourages advantage maximization. Unlike conventional policy gradient methods that rely on explicit value function approximation \cite{schulman2015high}, GRPO directly optimizes advantage by standardizing reward scores across samples. A distinction lies in GVPO's omission of standard deviation normalization.\footnote{\kc{This omission is not a heuristic design but rather a consequence of the theoretical formulation.}} Prior research \cite{liu2025understandingr1zeroliketrainingcritical} has also demonstrated that such scaling introduces bias by conflating prompt-level difficulty with reward signals.

\item the term \(Cov(\log\pi_\theta, \log\pi_{\theta^\prime})\) serves to constrain deviations of the policy \(\pi_\theta\) from policy $\pi_{\theta^\prime}$, corresponding to \(\mathbb{D}_{\text{KL}}[\pi_\theta||\pi_{\theta^\prime}]\). Moreover, in GVPO, where $\pi_{\theta^\prime}=\pi_{\theta_{\text{old}}}$, this term essentially aligns with the trust-region constraint \cite{pmlr-v37-schulman15}, that ensures robustness between policy updates.

\item the term \(Var(\log\pi_\theta)\) strikes a balance between exploration and exploitation. We juxtapose this term with entropy regularization \(-\mathbb{E}_y\log\pi(y|x)\) \cite{ahmed2019understanding}. 
\begin{itemize}[leftmargin=0.5cm]
\item Increasing entropy encourages diversity by driving the policy toward a uniform distribution, but risks suppressing the likelihood of high-quality responses. Conversely, reducing entropy concentrates probability mass on a narrow set of outputs, diminishing diversity and potentially inducing entropy collapse. Consequently, entropy regularization proves highly sensitive to its coefficient, complicating practical implementation.
\item In contrast, \(Var(\log\pi_\theta)\) circumvents this issue without requiring ad-hoc tuning by enabling scenarios where \kc{some responses receive zero probability, while other responses retain comparable probabilities. As the term $R - \overline{R}$ maximizes advantage, undesirable responses will receive zero probability, while favorable responses will maintain similar probabilities.
\item To illustrate the benefit, consider a toy example involving generation over the tokens $a, b, c, d, e$, where $\pi_\theta(a) = \pi_\theta(b) = 0$ and $\pi_\theta(c) = \pi_\theta(d) = \pi_\theta(e) = 1/3$. In this case, $Var(\log(\pi_\theta)) = 0$, indicating minimum variance and thus no penalty on this distribution. In contrast, entropy regularization only reaches its minimum either when the distribution is uniform or when it is one-hot, depending on whether one encourages higher or lower entropy.

}
\end{itemize}
\end{itemize}

%% file: sections/4.6.TRPO.tex
\kc{\textbf{In-depth similarities}. In addition to structural similarities}, we now highlight their key theoretical and practical distinctions. Modern policy gradient methods \cite{pmlr-v37-schulman15,schulman2017proximal,shao2024deepseekmath} optimize the expected reward under the current policy $\pi_\theta$ while constraining updates to avoid excessive deviation from the previous policy $\pi_{\theta_{\text{old}}}$. This is typically achieved by optimizing a  objective that combines the reward $R(x,y)$ and a KL-divergence penalty term $\mathbb{D}_{\text{KL}}[\pi_\theta||\pi_{\theta_{\text{old}}}]$, yielding the gradient expression:
\kc{
\begin{equation} 
\begin{aligned}
\label{equation:pg}
\nabla_\theta[\mathbb{E}_{x,y\sim\pi_\theta(y|x)}[R(x,y)]-\mathbb{D}_{\text{KL}}[\pi_\theta||\pi_{\theta_{\text{old}}}]] 
= \nabla_\theta\mathbb{E}_{x} \sum_y \pi_\theta(y|x)(R(x,y)-\log\frac{\pi_\theta(y|x)}{\pi_{\theta_{\text{old}}}(y|x)})& \\
= \mathbb{E}_{x,y\sim\pi_\theta(y|x)}(R(x,y)-\log\frac{\pi_\theta(y|x)}{\pi_{\theta_{\text{old}}}(y|x)}-1)  \nabla_\theta \log \pi_\theta(y|x)&
\end{aligned}
\end{equation}
}
However, estimating this expectation requires on-policy sampling from $\pi_\theta(y|x)$, leading to low sample efficiency---a well-documented limitation of policy gradient methods. Reusing stale samples from prior policies introduces bias, degrading optimization stability and final performance.  

To mitigate this, prior works~\cite{pmlr-v37-schulman15,schulman2017proximal,shao2024deepseekmath} employ importance sampling, rewriting Equation~\ref{equation:pg} as:
\begin{equation}
\mathbb{E}_{x,y\sim\pi_{\theta_{\text{old}}}(y|x)}\frac{\pi_\theta(y|x)}{\pi_{\theta_{\text{old}}}(y|x)}\left(R(x,y)-\log\frac{\pi_\theta(y|x)}{\pi_{\theta_{\text{old}}}(y|x)} -1 \right) \nabla_\theta \log \pi_\theta(y|x).
\end{equation}
This allows off-policy gradient estimation using samples from $\pi_{\theta_{\text{old}}}$. However, $\frac{\pi_\theta(y|x)}{\pi_{\theta_{\text{old}}}(y|x)}$ becomes unstable when $\pi_\theta$ deviates significantly from $\pi_{\theta_{\text{old}}}$, risking gradient explosion. Heuristics like gradient clipping \cite{schulman2017proximal} address this at the cost of biased gradient estimates, undermining theoretical guarantees.  

GVPO circumvents these issues because it does not necessitate on-policy sampling in the first place. By rearranging Equation~\ref{equation:pg}, we observe that the policy gradient can be expressed as:
\begin{equation}\nonumber
\mathbb{E}_{x,y\sim\pi_\theta(y|x)}\left[R(x,y)-\log\frac{\pi_\theta(y|x)}{\pi_{\theta_{\text{old}}}(y|x)} - \mathbb{E}_{y\sim\pi_\theta(y|x)}\left(R(x,y)-\log\frac{\pi_\theta(y|x)}{\pi_{\theta_{\text{old}}}(y|x)}\right)\right] \nabla_\theta \log \pi_\theta(y|x),
\end{equation}
where the baseline term (subtracted expectation) arises because $\mathbb{E}_{y\sim\pi_\theta} [c \nabla_\theta \log \pi_\theta(y|x)] = \nabla_\theta c = 0$ for any constant $c$. Crucially, the GVPO gradient generalizes this structure, $\nabla_\theta\hat{\mathcal{L}}_{\text{GVPO}}(\theta) = $
\begin{equation}\nonumber
\mathbb{E}_{x,y\sim\pi_s(y|x)}\left[R(x,y)-\log\frac{\pi_\theta(y|x)}{\pi_{\theta_{\text{old}}}(y|x)} - \mathbb{E}_{y\sim\pi_s(y|x)}\left(R(x,y)-\log\frac{\pi_\theta(y|x)}{\pi_{\theta_{\text{old}}}(y|x)}\right)\right] \nabla_\theta \log \pi_\theta(y|x),
\end{equation}
This reveals that classical policy gradient under trust-region constraint is a special case of GVPO gradient with $\pi_s = \pi_\theta$. As proven in Theorem~\ref{theorem:gvpo} and Theorem~\ref{corollary:pi_s}, GVPO retains the same optimal solution as the policy gradient method while decoupling the sampling distribution $\pi_s$ from the learned policy $\pi_\theta$. GVPO’s decoupling addresses two critical limitations:  
\begin{enumerate} [leftmargin=0.5cm]
    \item \textbf{Sample Efficiency}: Unlike on-policy methods \cite{singh2000convergence, williams1992simple}, GVPO supports off-policy training with reusable or mixed data (e.g., expert demonstrations, historical policies, or model distillations).  
    \item \textbf{Stability}: By avoiding importance sampling weights $\frac{\pi_\theta}{\pi_{\theta_{\text{old}}}}$, GVPO eliminates gradient explosion risks without biased clipping.  
\end{enumerate}
By synergizing these advantages, GVPO emerges as a competitive online reinforcement learning algorithm capable of leveraging diverse data sources, sustaining stable policy updates, and preserving convergence to optimality—a combination previously unattained in prior policy gradient methods.

%% file: sections/6.experiment.tex
\input{sections/tables/table_main}
\section{Experiments}
\vspace{-2.5mm}

\textbf{Task.} Following the established experimental setting of GRPO, we conduct a comprehensive evaluation on math reasoning. Specifically, we post-train the Qwen2.5-Math-7B model on Competition Math dataset \cite{hendrycksmath2021} and assess performance on AIME2024 \cite{li2024numinamath}, AMC \cite{li2024numinamath}, Math500 \cite{hendrycks2021measuring}, Minerva \cite{lewkowycz2022solving}, and OlympiadBench \cite{he2024olympiadbench}. For answer verification, we utilize the xVerify framework \cite{chen2025xverifyefficientanswerverifier}. We adopt the $pass@1$ accuracy for all benchmarks except AIME2024, where we report $avg@32$ accuracy to account for its limited size (30 problems) and high difficulty. 

\kc{In addition to math reasoning, we also evaluate GVPO on the summarization task in Appendix ~\ref{sec:sum}.}

\textbf{Setup.} To ensure a fair comparison across methods, we maintain identical experimental settings while only modifying the algorithmic component. For GVPO, we employ $\beta=0.1$ and $\pi_s=\pi_{\theta_{\text{old}}}$ in the main experiment. For competing approaches, we utilize hyperparameters specified in their original publications. All experiments generate $k=5$ responses per prompt. A comprehensive description of the training details is provided in Appendix~\ref{app:exp_setting}.

\textbf{Main Result.} Table~\ref{tab:results} shows the main experiment result, which demonstrates that GVPO achieves the best performance, outperforming both the base model and other variants in all benchmarks \cite{liu2025understanding, li2023remax, hu2025reinforce++}, particularly in complex problem-solving scenarios.
We attribute its effectiveness to its strong theoretical guarantees of convergence.

\textbf{Ablation on $\beta$.}  Figure~\ref{fig:beta} analyzes the sensitivity of GVPO to variations in $\beta$. The results demonstrate little fluctuation in performance across $\beta$, suggesting GVPO exhibits robustness to this hyperparameter. This stability may reduce the need for exhaustive tuning and enhance its practical utility.

\textbf{Ablation on $k$.} Figure~\ref{fig:k} examines how GVPO scales with $k$, evaluated on Qwen2.5-Math-1.5B. Top and bottom panels show results for MATH500 and AIME2024 respectively. GVPO consistently outperforms GRPO across all $k$ and demonstrates superior scalability. Notably, GVPO matches the AIME2024 performance of a 7B model on the 1.5B architecture through increased $k$, highlighting its potential for reducing inference costs in practice.

\textbf{Ablation on $\pi_s$.} Figure~\ref{fig:pi_s} investigates GVPO's versatility on sampling distributions, evaluated on Qwen2.5-Math-1.5B and MATH500. We propose a heuristic $\pi_s$ that mixes responses from $\pi_{\theta_{\text{old}}}$ with historical responses. Results demonstrate GVPO’s robust performance across mixing proportions, highlighting: (1) this $\pi_s$ can reduce sampling costs during training, and (2) it suggests GVPO’s potential to bridge modern LLM research with previous RL research on exploration strategies.

\vspace{-5mm}
\begin{minipage}[b]{0.32\textwidth}
  \begin{figure}[H] 
    \centering
    \input{sections/tables/plot_beta}
    \vspace{-7mm}
    \caption{Ablation on $\beta$. Each line represents a dataset.}
    \label{fig:beta}
  \end{figure}
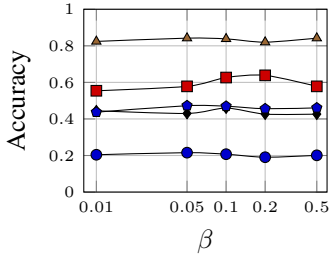
\end{minipage}
\hfill 
\begin{minipage}[b]{0.32\textwidth}
  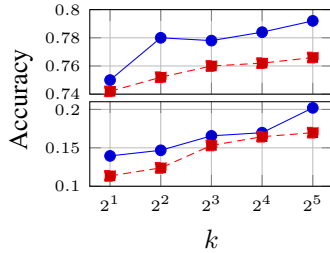
\begin{figure}[H]
    \centering
    \input{sections/tables/plot_k}
    \vspace{-7mm}
    \caption{Ablation on $k$. Blue line: \textcolor{blue}{GVPO}; Red line: \textcolor{red}{GRPO}.}
    \label{fig:k}
  \end{figure}
\end{minipage}
\hfill
\begin{minipage}[b]{0.32\textwidth}
  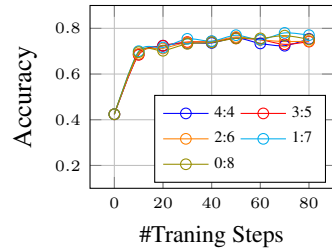
\begin{figure}[H]
    \centering
    \input{sections/tables/plot_pi_s_ii}
    \vspace{-7mm}
    \caption{Ablation on $\pi_s$. \#(historical $y$) : \#($y$ from $\pi_{\theta_{\text{old}}}$)}
    \label{fig:pi_s}
  \end{figure}
\end{minipage}

\kc{
\input{sections/tables/table_reg}

\textbf{Ablation on Regularization Terms.} In Section~\ref{sec:reg}, we decompose the GVPO loss into three components: an advantage maximization term, a variance regularization term ($Var$) on the current policy, and a covariance regularization term ($Cov$) between the current and reference policies. Table~\ref{tab:reg} presents the ablation results for these regularization terms. We first remove $Var(\log\pi_\theta)$ and $Cov(\log\pi_\theta, \log\pi_{\theta^\prime})$ individually. In both cases, the model fails to converge and generates incoherent outputs, indicating that each term plays a crucial role in stabilizing training. When both regularization terms are removed simultaneously—reducing the objective to $R - \overline{R}$—the model initially converges but diverges after approximately 10\% of the training steps, further confirming that these regularization components are essential to GVPO’s stability.

\textbf{Study on $Var(\log\pi_\theta)$.} We further investigate the role of $Var(\log\pi_\theta)$ by replacing it with an entropy regularization term. As shown in Table~\ref{tab:reg}, the model again fails to converge and produces incoherent outputs. Lowering the learning rate to $1\text{e}{-6}$ improves stability marginally, yet substituting it with entropy regularization still lead to divergence or suboptimal performance. These findings suggest that although entropy regularization serves a similar stabilizing purpose, it cannot fully substitute $Var(\log\pi_\theta)$. The degradation likely arises from entropy regularization being either too weak—insufficient to suppress extreme updates—or too strong—impeding convergence. This underscores a key limitation of entropy regularization: its sensitivity to coefficient tuning. In contrast, the coefficient for $Var(\log\pi_\theta)$ in GVPO is derived analytically via the optimal solution theorem, eliminating the need for manual tuning and enhancing overall robustness.

\input{sections/tables/table_seed}

\textbf{Robustness Check with Different Random Seeds.} We conducted additional experiments using 10 random seeds for both methods on Qwen2.5-Math-1.5B. The results show that GVPO consistently outperforms GRPO in overall performance while exhibiting comparable standard deviations, indicating stable results across different runs.

\input{sections/tables/table_base}

\textbf{Robustness Check with a Different Foundation Model.} To further assess generalization, we repeated the experiments using Llama-3.1-8B-Instruct as the foundation model. As shown in Table~\ref{tab:base}, GVPO again outperforms GRPO, reaffirming the robustness and effectiveness of our approach.

}

%% file: sections/tables/table_main.tex
\begin{table}[htbp]
\centering
\caption{Algorithm Performance Comparison on Mathematical Datasets}
\label{tab:results}
\begin{tabular}{lccccc}
\toprule
Algorithm & AIME2024 & AMC & MATH500 & Minerva & Olympiadbench \\
\midrule
Qwen2.5-Math-7B & 14.68 & 38.55 & 64.00 & 27.20 & 30.66  \\
\quad+GRPO     & 14.79 & 55.42 & 80.00 & 41.17 & 42.07  \\
\quad+Dr.GRPO  & 16.56 & 48.19 & 81.20 & 44.48 & 43.40  \\
\quad+Remax  & 17.19 & 60.24 & 82.00 & 40.44 & 45.19  \\
\quad+Reinforce++  & 16.67 & 54.22 & 80.40 & 43.01 & 41.78  \\
\quad+\textbf{GVPO}     & \textbf{20.72} & \textbf{62.65} & \textbf{83.80} & \textbf{45.95} & \textbf{46.96} \\
\bottomrule
\end{tabular}
\vspace{-4.5mm}
\end{table}

%% file: sections/tables/plot_beta.tex
\begin{tikzpicture}
\begin{axis}[
    xlabel=$\beta$,
    ylabel=Accuracy,
    xmode=log,
    log ticks with fixed point,
    xtick={0.01,0.05,0.1,0.2,0.5},
    xmin=0.008, xmax=0.6,
    ymin=0, ymax=1.0,
    grid=both,
    legend style={},
    ticklabel style = {font=\tiny},
    width=4.8cm,
    height=4cm,
]

\addplot+[smooth, black, mark=*] coordinates {
    (0.01, 0.2041)
    (0.05, 0.2156)
    (0.1, 0.2072)
    (0.2, 0.1906)
    (0.5, 0.2010)
};

\addplot+[smooth, black, mark=square*] coordinates {
    (0.01, 0.5542)
    (0.05, 0.5783)
    (0.1, 0.6265)
    (0.2, 0.6385)
    (0.5, 0.5783)
};

\addplot+[smooth, black, mark=triangle*] coordinates {
    (0.01, 0.8240)
    (0.05, 0.8420)
    (0.1, 0.8380)
    (0.2, 0.82)
    (0.5, 0.8420)
};

\addplot+[smooth, black, mark=diamond*] coordinates {
    (0.01, 0.4448)
    (0.05, 0.4301)
    (0.1, 0.4595)
    (0.2, 0.4264)
    (0.5, 0.4264)
};

\addplot+[smooth, black, mark=pentagon*] coordinates {
    (0.01, 0.4370)
    (0.05, 0.4725)
    (0.1, 0.4696)
    (0.2, 0.4559)
    (0.5, 0.4607)
};

\end{axis}
\end{tikzpicture}

%% file: sections/tables/plot_k.tex
\begin{tikzpicture}
  \begin{groupplot}[
    group style={
      group size=1 by 2,
      vertical sep=0.1cm,
      group name=myplots,
      x descriptions at=edge bottom  
    },
    xmode=log,
    log basis x=2,
    width=4.8cm, 
    height=2.7cm,
    grid=major,
    xtick={2,4,8,16,32},
    legend columns=2,
    legend style={
      at={([yshift=-15pt]myplots c1r2.south)},
      anchor=north
    },
    ticklabel style = {font=\tiny},
    legend to name=commonlegend
    ]
    
    \nextgroupplot[
      xlabel=,             
      ymin=0.74, ymax=0.8  
    ]
    \addplot+[solid, blue, mark=*] coordinates {
      (2, 0.7500) (4, 0.78) (8, 0.778) (16, 0.784) (32, 0.792)
    };
    \addlegendentry{GVPO}
    
    \addplot+[densely dashed, red, mark=square*] coordinates {
      (2, 0.7420) (4, 0.7520) (8, 0.7600) (16, 0.7620) (32, 0.7660)
    };
    \addlegendentry{GRPO}

    \nextgroupplot[
      ymin=0.1, ymax=0.21  
      ,xlabel=$k$
    ]
    \addplot+[solid, blue, mark=*] coordinates {
      (2, 0.1395) (4, 0.1468) (8, 0.1656) (16, 0.1697) (32, 0.202)
    };
    
    \addplot+[densely dashed, red, mark=square*] coordinates {
      (2, 0.1135) (4, 0.1239) (8, 0.1531) (16, 0.1645) (32, 0.1697)
    };

  \end{groupplot}

  \node[rotate=90, anchor=center] at ([xshift=-2mm,yshift=-6mm]myplots c1r1.outer west) {Accuracy};
  
  \node[anchor=north] at (myplots c1r2.south) {\pgfplotslegendfromname{commonlegend}};
\end{tikzpicture}

%% file: sections/tables/plot_pi_s_ii.tex
\begin{tikzpicture}
\begin{axis}[
    xlabel=\#Traning Steps,
    xlabel style = {font=\small},
    ylabel=Accuracy,
    log ticks with fixed point,
    xtick={0,20,40,60,80},
    xmin=-10, xmax=90,
    ymin=0.1, ymax=0.9,
    grid=both,
    legend style={font=\tiny},
    ticklabel style = {font=\tiny},
    width=4.8cm,
    height=4cm,
    legend pos = south east,
    legend columns=2, 
]

\addplot+[smooth, blue, mark=o] coordinates {
(0, 0.424)
(10, 0.686)
(20, 0.724)
(30, 0.736)
(40, 0.736)
(50, 0.758)
(60, 0.734)
(70, 0.724)
(80, 0.754)
};
\addlegendentry{4:4}

\addplot+[smooth, red, mark=o] coordinates {
(0, 0.424)
(10, 0.684)
(20, 0.724)
(30, 0.74)
(40, 0.742)
(50, 0.756)
(60, 0.75)
(70, 0.73)
(80, 0.742)
};
\addlegendentry{3:5}

\addplot+[smooth, orange, mark=o] coordinates {
(0, 0.424)
(10, 0.688)
(20, 0.712)
(30, 0.738)
(40, 0.742)
(50, 0.762)
(60, 0.75)
(70, 0.742)
(80, 0.744)
};
\addlegendentry{2:6}

\addplot+[smooth, cyan, mark=o] coordinates {
(0, 0.424)
(10, 0.7)
(20, 0.716)
(30, 0.754)
(40, 0.744)
(50, 0.77)
(60, 0.75)
(70, 0.78)
(80, 0.77)
};
\addlegendentry{1:7}

\addplot+[smooth, olive, mark=o] coordinates {
(0, 0.424)
(10, 0.696)
(20, 0.702)
(30, 0.732)
(40, 0.738)
(50, 0.756)
(60, 0.756)
(70, 0.768)
(80, 0.752)
};
\addlegendentry{0:8}

\end{axis}
\end{tikzpicture}

%% file: sections/tables/table_reg.tex
\vspace{-4mm}
\begin{table}[htbp]
\centering
\caption{Ablation on Regularization Terms}
\label{tab:reg}
\begin{tabular}{lccccc}
\toprule
Algorithm & AIME24 & AMC & MATH500 & Minerva & Olympiadbench \\
\midrule
GVPO     & 20.72 & 62.65 & 83.80 & 45.95 & 46.96  \\
GVPO - Var     & 0.00 & 0.00 & 0.00 & 0.00 & 0.00  \\
GVPO - Cov     & 0.00 & 0.00 & 0.00 & 0.00 & 0.00  \\
GVPO - Var - Cov     & 7.19 & 39.76 & 73.00 & 34.93 & 35.70  \\
\midrule
GVPO - Var + Entropy     & 0.00 & 0.00 & 0.00 & 0.00 & 0.00  \\
GVPO - Var + Entropy (LR=1e-6)& 3.02 & 39.76 & 33.20 & 23.16 & 8.89  \\
\bottomrule
\end{tabular}
\end{table}

%% file: sections/tables/table_seed.tex
\vspace{-4mm}
\begin{table}[htbp]
\centering
\caption{Algorithm Performance Comparison with Different Random Seeds}
\label{tab:base}
\begin{tabular}{lccccc}
\toprule
Algorithm & AIME2024 & AMC & MATH500 & Minerva & Olympiadbench \\
\midrule
GRPO     & 13.59$\pm$1.11 & 44.34$\pm$2.19 & 76.30$\pm$1.35 & 35.40$\pm$1.04 & 38.65$\pm$0.74  \\
\textbf{GVPO}     & \textbf{15.56$\pm$1.05} & \textbf{45.78$\pm$2.05} & \textbf{77.36$\pm$0.98} & \textbf{36.03$\pm$1.15} & \textbf{39.58$\pm$1.08} \\
\bottomrule
\end{tabular}
\end{table}

%% file: sections/tables/table_base.tex
\vspace{-4mm}
\begin{table}[htbp]
\centering
\caption{Algorithm Performance Comparison with Llama-3.1-8B-Instruct}
\label{tab:base}
\begin{tabular}{lccccc}
\toprule
Algorithm & AIME2024 & AMC & MATH500 & Minerva & Olympiadbench \\
\midrule
Llama-3.1-8B-Instruct & 5.00 & 19.27 & 50.40 & 22.42 & 17.04  \\
\quad+GRPO     & 8.54 & 19.27 & 52.60 & 25.37 & 16.15  \\
\quad+\textbf{GVPO}     & \textbf{11.56} & \textbf{20.48} & \textbf{56.60} & \textbf{29.41} & \textbf{20.59} \\
\bottomrule
\end{tabular}
\end{table}

%% file: sections/2.related.tex
\kc{
\section{Related Work}

This paper closely relates to the LLM post-training literature, including SFT \cite{ouyang2022training, fan2025preference}, RLHF \cite{bai2022training}, PPO \cite{schulman2017proximal}, TRPO \cite{pmlr-v37-schulman15}, DPO \cite{rafailov2023direct}, GRPO \cite{shao2024deepseekmath}, Dr.GRPO \cite{liu2025understanding}, Remax \cite{li2023remax}, Reinforce++ \cite{hu2025reinforce++}, DAPO \cite{yu2025dapo}; and debates on whether RL improves LLM reasoning capacity \cite{yue2025does}, spurious reward issue \cite{shao2025spurious}, data contamination issue \cite{wu2025reasoning}, and Pass@K optimization \cite{zhang2025rspo}.

Beyond post-training, LLMs have demonstrated broad applicability across domains, including vision-language modeling \cite{guo2025logic}, graph learning \cite{jiang2024killing, hu2024let}, AI safety research \cite{xu2025mark}, evaluation frameworks leveraging LLM-as-a-Judge paradigms \cite{hu2024language, hu2024explaining}, retrieval-augmented generation \cite{dai2025seper}, AI for Education \cite{hu2025unveiling, wang2025llm} and economics of AI \cite{zhang2023impact, zhang2024optimized}. These broader developments underscore the promising role of post-training in enhancing the adaptability of LLMs across a growing range of applications.
}

%% file: sections/7.conclusions.tex
\section{Conclusion}

\vspace{-2mm}

In this paper, we present Group Variance Policy Optimization (GVPO). GVPO guarantees a unique optimal solution, exactly the KL-constrained reward maximization objective. Moreover, it supports flexible sampling distributions that avoids on-policy and importance sampling limitations. Through systematic comparisons with other prominent methods both theoretically and empirically, we establish GVPO as a new paradigm for reliable and versatile LLM post-training.

%% file: sections/90.appendix.tex
\appendix
\section*{Appendix}
\section{Supplementary Experiment Information}
\subsection{Experiment Settings}
\label{app:exp_setting}
\textbf{Hyperparameter Recipe.} For each step, we sample $1024$ prompts from the training set and set the mini-batch size in each step to $256$. We repeat the whole training set for 10 epochs and set the warm-up ratio to $5\%$. We grid-search the learning rate in $\{5e-7,\ 1e-6, 5e-6, 1e-5\}$ and find $5e-6$ to be the best setting. 
We conduct the main experiment using an Deepseek-R1-like chat template on top of Qwen2.5-Math-7B as in \cite{hu2025openreasonerzeroopensourceapproach}. In the ablation experiments, for faster training and GPU memory limitations, we use the original Qwen chat template on top of Qwen2.5-Math-1.5B.

\textbf{Compute Resources.} We conduct our experiments using a server with eight 80GB H800 GPU cards. For Qwen2.5-Math-7B experiments with $k=5$, it takes 6 to 8 minutes per training step and approximately 12 hours per experiment. For Qwen2.5-Math-1.5B experiments with $k=8$, it takes 4 to 5 minutes per training step and approximately 8 hours per experiment.

\subsection{Code Implementation}

\label{app:code}
It is easy to implement GVPO based on open-source RL framework. For example\footnote{Make sure that 1) each input batch correspond to the $k$ responses of a prompt and 2) the std normalizer in the GRPO advantage calculation has been removed.}, we show the minimum viable implementation of GVPO that only modifies a \href{https://github.com/volcengine/verl/blob/main/verl/trainer/ppo/core_algos.py#L354}{few line}  of GRPO loss in verl \cite{sheng2024hybridflow}:
\input{sections/tables/code_GVPO}

\kc{
Moreover, we provide an implementation based on verl at https://github.com/jszkc/GVPO.
}

\input{sections/91.proofs}

\section{Alternative Task on Summarization} \label{sec:sum}
To evaluate GVPO in the context of alignment \footnote{Thank Reviewer X8wt for sponsoring this study.}, we conduct experiments on a summarization task following the experimental setup of DPO \cite{rafailov2023direct}. Specifically, we used the Reddit dataset from \cite{stiennon2020learning} and the reward model OpenAssistant/reward-model-deberta-v3-large-v2. First, we fine-tuned Qwen2.5 1.5B using the dataset to obtain an SFT reference model. We then applied both DPO and GVPO for post-training. The training data was sampled from the reference model and scored using the reward model.

For evaluation, we considered:
\begin{itemize}[leftmargin=0.5cm]
    \item The average reward of generated summaries on the test split.
    \item Win rate against human-written demonstration answers, as scored by the reward model.
    \item Preference accuracy using human-labeled comparison data: both DPO and GVPO are trained via
$R_\theta(x, y) = \beta \frac{\log \pi_\theta(y|x)}{\log \pi_{\text{ref}}(y|x)} + \beta \log Z(x)$. Given a preference pair $(y_w, y_l)$ where $y_w$ is preferred, we compute whether
$\frac{\log \pi_\theta(y_w|x)}{\log \pi_{\text{ref}}(y_w|x)} > \frac{\log \pi_\theta(y_l|x)}{\log \pi_{\text{ref}}(y_l|x)}$.
    \item Benchmarking by a set of powerful LLMs.
    \item Human evaluations by crowd-sourcing workers. We hire three workers to label their preference on 100 prompts on Prolific platform.
\end{itemize}

\begin{table}[htbp]
\centering
\caption{Algorithm Performance Comparison on Summarization}
\label{tab:base}
\begin{tabular}{lccccccc}
\toprule
Algorithm & Reward & Win Rate & Accuracy & Gpt-4o & Gemini-2.5-pro & Deepseek-R1 & Human \\
\midrule
SFT & 2.84 & 41.85\% & --- & --- & --- & --- & ---  \\
+DPO & 4.83 & 68.28\% & 60.43\% & 31\% & 40\% & 25\% & 34\%  \\
+GVPO & 5.75 & 79.49\% & 64.93\% & 69\% & 60\% & 75\% & 66\% \\
\bottomrule
\end{tabular}
\end{table}

The results indicate that GVPO outperforms DPO across these metrics, supporting the view that improvements in policy search methods are positively correlated with alignment quality.

%% file: sections/tables/code_GVPO.tex
\begin{lstlisting}[caption={A Simple GVPO Code Implementation}, label={lst:policy_loss}]
def compute_policy_loss(old_log_prob, log_prob, advantages, eos_mask, **kwargs):
    scores = (log_prob * eos_mask).sum(dim=-1)
    scores_old = (old_log_prob * eos_mask).sum(dim=-1)
    advs = (advantages * eos_mask).sum(dim=-1) / eos_mask.sum(dim=-1)
    
    beta = 0.1
    k = scores.size(0)
    
    scores_new = scores.detach()
    loss = -1 * beta * scores * (advs - beta * ((scores_new - scores_new.mean()) - (scores_old - scores_old.mean())))
    
    return loss.sum()/(k-1)
\end{lstlisting}

%% file: sections/91.proofs.tex
\section{Proofs}
\label{app:proof}
\subsection{Proof of Theorem~\ref{theorem:gvpo}}
\label{app:proof1}
\textbf{Theorem~\ref{theorem:gvpo}}. \textit{The unique optimal policy that minimizes $\hat{\mathcal{L}}_{\text{GVPO}}(\theta)$, defined as
\[
\hat{\mathcal{L}}_{\text{GVPO}}(\theta)=
\mathbb{E}_{x \sim \mathcal{D}} \mathbb{E}_{y \sim \pi_{s}(\cdot|x)}[(R_\theta(x,y)-
\mathbb{E}_{y \sim \pi_{s}}R_\theta(x,y))-(R(x,y)-
\mathbb{E}_{y \sim \pi_{s}}R(x,y))]^2
\]
, is given by $\pi_\theta (y|x)=\pi^* (y|x)= \frac{1}{Z(x)}\pi_{\theta^\prime}(y|x)e^{R(x,y)/\beta}$ 
for $\pi_s=\pi_{\theta^\prime}$.}

\begin{proof}
We prove the theorem by establishing both necessity and sufficiency.

\textbf{Necessity:} If $\pi_\theta(y|x) = \pi^*(y|x)$, then it is an optimal policy solution.  

The loss function $\hat{\mathcal{L}}_{\text{GVPO}}(\theta)$ is non-negative because it represents the expectation of a squared term:
\[
\hat{\mathcal{L}}_{\text{GVPO}}(\theta) = \mathbb{E}_{x,y} \left[ \big((R_\theta(x,y) - \mathbb{E}_y R_\theta(x,y)) - (R(x,y) - \mathbb{E}_y R(x,y)\big)^2 \right] \geq 0.
\]
When $\pi_\theta(y|x) = \pi^*(y|x)$, we have $R_\theta(x,y) = R(x,y)$. Substituting this into the loss function gives $\hat{\mathcal{L}}_{\text{GVPO}}(\theta) = 0$, confirming that $\pi^*$ achieves the minimum loss.

\textbf{Sufficiency:} If a policy $\pi_\theta$ is optimal, then $\pi_\theta(y|x) = \pi^*(y|x)$.  

Assume for contradiction that there exists an optimal policy $\pi_\theta \neq \pi^*$. Since $\pi_\theta$ is optimal, $\hat{\mathcal{L}}_{\text{GVPO}}(\theta) = 0$. This implies:
\[
(R_\theta(x,y) - \mathbb{E}_y R_\theta(x,y)) = (R(x,y) - \mathbb{E}_y R(x,y)), \quad \forall x, y \text{ s.t. } \pi_s(y|x)>0
\]
Rewriting $R_\theta$ and $R$ in terms of their respective policies:
\[
\beta log \pi_\theta(y|x) - \mathbb{E}_y R_\theta(x,y) = \beta log \pi^*(y|x) - \mathbb{E}_y R(x,y).
\]
Rearranging terms yields:
\[
\pi_\theta(y|x) = \exp\left(\frac{\mathbb{E}_y [R_\theta(x,y) - R(x,y)]}{\beta}\right) \pi^*(y|x).
\]
Since $\sum_{y\in\{y|\pi_{\theta^\prime}(y|x)>0\}} \pi_\theta(y|x) = \sum_{y\in\{y|\pi_{\theta^\prime}(y|x)>0\}} \pi^*(y|x) = 1$, we must have:
\[
\sum_y\pi_\theta(y|x) = \exp\left(\frac{\mathbb{E}_y [R_\theta(x,y) - R(x,y)]}{\beta}\right) \sum_y\pi^*(y|x)
\]
\[
\implies \exp\left(\frac{\mathbb{E}_y [R_\theta(x,y) - R(x,y)]}{\beta}\right) = 1
\]
Thus, $\pi_\theta(y|x) = \pi^*(y|x)$ for all $x, y$, contradicting the assumption $\pi_\theta \neq \pi^*$.  

Since both necessity and sufficiency hold, the optimal policy is uniquely $\pi^*$.
\end{proof}

%% file: main.bib
@misc{hu2025openreasonerzeroopensourceapproach,
      title={Open-Reasoner-Zero: An Open Source Approach to Scaling Up Reinforcement Learning on the Base Model}, 
      author={Jingcheng Hu and Yinmin Zhang and Qi Han and Daxin Jiang and Xiangyu Zhang and Heung-Yeung Shum},
      year={2025},
      eprint={2503.24290},
      archivePrefix={arXiv},
      primaryClass={cs.LG},
      url={https://arxiv.org/abs/2503.24290}, 
}

@article{sheng2024hybridflow,
  title   = {HybridFlow: A Flexible and Efficient RLHF Framework},
  author  = {Guangming Sheng and Chi Zhang and Zilingfeng Ye and Xibin Wu and Wang Zhang and Ru Zhang and Yanghua Peng and Haibin Lin and Chuan Wu},
  year    = {2024},
  journal = {arXiv preprint arXiv: 2409.19256}
}

@misc{chen2025xverifyefficientanswerverifier,
      title={xVerify: Efficient Answer Verifier for Reasoning Model Evaluations}, 
      author={Ding Chen and Qingchen Yu and Pengyuan Wang and Wentao Zhang and Bo Tang and Feiyu Xiong and Xinchi Li and Minchuan Yang and Zhiyu Li},
      year={2025},
      eprint={2504.10481},
      archivePrefix={arXiv},
      primaryClass={cs.CL},
      url={https://arxiv.org/abs/2504.10481}, 
}

@InProceedings{pmlr-v37-schulman15,
  title = 	 {Trust Region Policy Optimization},
  author = 	 {Schulman, John and Levine, Sergey and Abbeel, Pieter and Jordan, Michael and Moritz, Philipp},
  booktitle = 	 {Proceedings of the 32nd International Conference on Machine Learning},
  pages = 	 {1889--1897},
  year = 	 {2015},
  editor = 	 {Bach, Francis and Blei, David},
  volume = 	 {37},
  series = 	 {Proceedings of Machine Learning Research},
  address = 	 {Lille, France},
  month = 	 {07--09 Jul},
  publisher =    {PMLR},
  url = 	 {https://proceedings.mlr.press/v37/schulman15.html}
}

@misc{zhao2025surveylargelanguagemodels,
      title={A Survey of Large Language Models}, 
      author={Wayne Xin Zhao and Kun Zhou and Junyi Li and Tianyi Tang and Xiaolei Wang and Yupeng Hou and Yingqian Min and Beichen Zhang and Junjie Zhang and Zican Dong and Yifan Du and Chen Yang and Yushuo Chen and Zhipeng Chen and Jinhao Jiang and Ruiyang Ren and Yifan Li and Xinyu Tang and Zikang Liu and Peiyu Liu and Jian-Yun Nie and Ji-Rong Wen},
      year={2025},
      eprint={2303.18223},
      archivePrefix={arXiv},
      primaryClass={cs.CL},
      url={https://arxiv.org/abs/2303.18223}, 
}

@misc{minaee2025largelanguagemodelssurvey,
      title={Large Language Models: A Survey}, 
      author={Shervin Minaee and Tomas Mikolov and Narjes Nikzad and Meysam Chenaghlu and Richard Socher and Xavier Amatriain and Jianfeng Gao},
      year={2025},
      eprint={2402.06196},
      archivePrefix={arXiv},
      primaryClass={cs.CL},
      url={https://arxiv.org/abs/2402.06196}, 
}

@misc{tie2025surveyposttraininglargelanguage,
      title={A Survey on Post-training of Large Language Models}, 
      author={Guiyao Tie and Zeli Zhao and Dingjie Song and Fuyang Wei and Rong Zhou and Yurou Dai and Wen Yin and Zhejian Yang and Jiangyue Yan and Yao Su and Zhenhan Dai and Yifeng Xie and Yihan Cao and Lichao Sun and Pan Zhou and Lifang He and Hechang Chen and Yu Zhang and Qingsong Wen and Tianming Liu and Neil Zhenqiang Gong and Jiliang Tang and Caiming Xiong and Heng Ji and Philip S. Yu and Jianfeng Gao},
      year={2025},
      eprint={2503.06072},
      archivePrefix={arXiv},
      primaryClass={cs.CL},
      url={https://arxiv.org/abs/2503.06072}, 
}

@misc{zhou2023comprehensivesurveypretrainedfoundation,
      title={A Comprehensive Survey on Pretrained Foundation Models: A History from BERT to ChatGPT}, 
      author={Ce Zhou and Qian Li and Chen Li and Jun Yu and Yixin Liu and Guangjing Wang and Kai Zhang and Cheng Ji and Qiben Yan and Lifang He and Hao Peng and Jianxin Li and Jia Wu and Ziwei Liu and Pengtao Xie and Caiming Xiong and Jian Pei and Philip S. Yu and Lichao Sun},
      year={2023},
      eprint={2302.09419},
      archivePrefix={arXiv},
      primaryClass={cs.AI},
      url={https://arxiv.org/abs/2302.09419}, 
}

@article{ouyang2022training,
  title={Training language models to follow instructions with human feedback},
  author={Ouyang, Long and Wu, Jeffrey and Jiang, Xu and Almeida, Diogo and Wainwright, Carroll and Mishkin, Pamela and Zhang, Chong and Agarwal, Sandhini and Slama, Katarina and Ray, Alex and others},
  journal={Advances in neural information processing systems},
  volume={35},
  pages={27730--27744},
  year={2022}
}

@article{bai2022training,
  title={Training a helpful and harmless assistant with reinforcement learning from human feedback},
  author={Bai, Yuntao and Jones, Andy and Ndousse, Kamal and Askell, Amanda and Chen, Anna and DasSarma, Nova and Drain, Dawn and Fort, Stanislav and Ganguli, Deep and Henighan, Tom and others},
  journal={arXiv preprint arXiv:2204.05862},
  year={2022}
}

@article{sutton2019bitter,
  title={The bitter lesson},
  author={Sutton, Richard},
  journal={Incomplete Ideas (blog)},
  volume={13},
  number={1},
  pages={38},
  year={2019}
}

@article{shao2024deepseekmath,
  title={Deepseekmath: Pushing the limits of mathematical reasoning in open language models},
  author={Shao, Zhihong and Wang, Peiyi and Zhu, Qihao and Xu, Runxin and Song, Junxiao and Bi, Xiao and Zhang, Haowei and Zhang, Mingchuan and Li, YK and Wu, Y and others},
  journal={arXiv preprint arXiv:2402.03300},
  year={2024}
}

@article{schulman2017proximal,
  title={Proximal policy optimization algorithms},
  author={Schulman, John and Wolski, Filip and Dhariwal, Prafulla and Radford, Alec and Klimov, Oleg},
  journal={arXiv preprint arXiv:1707.06347},
  year={2017}
}

@article{guo2025deepseek,
  title={Deepseek-r1: Incentivizing reasoning capability in llms via reinforcement learning},
  author={Guo, Daya and Yang, Dejian and Zhang, Haowei and Song, Junxiao and Zhang, Ruoyu and Xu, Runxin and Zhu, Qihao and Ma, Shirong and Wang, Peiyi and Bi, Xiao and others},
  journal={arXiv preprint arXiv:2501.12948},
  year={2025}
}

@article{yu2025dapo,
  title={Dapo: An open-source llm reinforcement learning system at scale},
  author={Yu, Qiying and Zhang, Zheng and Zhu, Ruofei and Yuan, Yufeng and Zuo, Xiaochen and Yue, Yu and Fan, Tiantian and Liu, Gaohong and Liu, Lingjun and Liu, Xin and others},
  journal={arXiv preprint arXiv:2503.14476},
  year={2025}
}

@article{liu2025understanding,
  title={Understanding r1-zero-like training: A critical perspective},
  author={Liu, Zichen and Chen, Changyu and Li, Wenjun and Qi, Penghui and Pang, Tianyu and Du, Chao and Lee, Wee Sun and Lin, Min},
  journal={arXiv preprint arXiv:2503.20783},
  year={2025}
}

@article{gao2024towards,
  title={Towards a unified view of preference learning for large language models: A survey},
  author={Gao, Bofei and Song, Feifan and Miao, Yibo and Cai, Zefan and Yang, Zhe and Chen, Liang and Hu, Helan and Xu, Runxin and Dong, Qingxiu and Zheng, Ce and others},
  journal={arXiv preprint arXiv:2409.02795},
  year={2024}
}

@article{touvron2023llama,
  title={Llama 2: Open foundation and fine-tuned chat models},
  author={Touvron, Hugo and Martin, Louis and Stone, Kevin and Albert, Peter and Almahairi, Amjad and Babaei, Yasmine and Bashlykov, Nikolay and Batra, Soumya and Bhargava, Prajjwal and Bhosale, Shruti and others},
  journal={arXiv preprint arXiv:2307.09288},
  year={2023}
}

@article{rafailov2023direct,
  title={Direct preference optimization: Your language model is secretly a reward model},
  author={Rafailov, Rafael and Sharma, Archit and Mitchell, Eric and Manning, Christopher D and Ermon, Stefano and Finn, Chelsea},
  journal={Advances in Neural Information Processing Systems},
  volume={36},
  pages={53728--53741},
  year={2023}
}

@article{bradley1952rank,
  title={Rank analysis of incomplete block designs: I. The method of paired comparisons},
  author={Bradley, Ralph Allan and Terry, Milton E},
  journal={Biometrika},
  volume={39},
  number={3/4},
  pages={324--345},
  year={1952},
  publisher={JSTOR}
}

@inproceedings{jaques2017sequence,
  title={Sequence tutor: Conservative fine-tuning of sequence generation models with kl-control},
  author={Jaques, Natasha and Gu, Shixiang and Bahdanau, Dzmitry and Hern{\'a}ndez-Lobato, Jos{\'e} Miguel and Turner, Richard E and Eck, Douglas},
  booktitle={International Conference on Machine Learning},
  pages={1645--1654},
  year={2017},
  organization={PMLR}
}

@InProceedings{hong2024energybasedpreferencemodeloffers,
  title = 	 {Energy-Based Preference Model Offers Better Offline Alignment than the Bradley-Terry Preference Model},
  author =       {Hong, Yuzhong and Zhang, Hanshan and Bao, Junwei and Jiang, Hongfei and Song, Yang},
  booktitle = 	 {Proceedings of the 42nd International Conference on Machine Learning},
  pages = 	 {23787--23804},
  year = 	 {2025},
  volume = 	 {267},
  series = 	 {Proceedings of Machine Learning Research},
  month = 	 {13--19 Jul},
  publisher =    {PMLR},
  url = 	 {https://proceedings.mlr.press/v267/hong25i.html},
}

@article{tang2024generalized,
  title={Generalized preference optimization: A unified approach to offline alignment},
  author={Tang, Yunhao and Guo, Zhaohan Daniel and Zheng, Zeyu and Calandriello, Daniele and Munos, R{\'e}mi and Rowland, Mark and Richemond, Pierre Harvey and Valko, Michal and Pires, Bernardo {\'A}vila and Piot, Bilal},
  journal={arXiv preprint arXiv:2402.05749},
  year={2024}
}

@inproceedings{bong2022generalized,
  title={Generalized results for the existence and consistency of the MLE in the Bradley-Terry-Luce model},
  author={Bong, Heejong and Rinaldo, Alessandro},
  booktitle={International Conference on Machine Learning},
  pages={2160--2177},
  year={2022},
  organization={PMLR}
}

@article{sutton1999policy,
  title={Policy gradient methods for reinforcement learning with function approximation},
  author={Sutton, Richard S and McAllester, David and Singh, Satinder and Mansour, Yishay},
  journal={Advances in neural information processing systems},
  volume={12},
  year={1999}
}

@article{williams1992simple,
  title={Simple statistical gradient-following algorithms for connectionist reinforcement learning},
  author={Williams, Ronald J},
  journal={Machine learning},
  volume={8},
  pages={229--256},
  year={1992},
  publisher={Springer}
}

@inproceedings{ahmed2019understanding,
  title={Understanding the impact of entropy on policy optimization},
  author={Ahmed, Zafarali and Le Roux, Nicolas and Norouzi, Mohammad and Schuurmans, Dale},
  booktitle={International conference on machine learning},
  pages={151--160},
  year={2019},
  organization={PMLR}
}

@misc{wu2022asymptoticcomparisonidentifyingconstraints,
      title={Asymptotic comparison of identifying constraints for Bradley-Terry models}, 
      author={Weichen Wu and Brian W. Junker and Nynke M. D. Niezink},
      year={2022},
      eprint={2205.04341},
      archivePrefix={arXiv},
      primaryClass={math.ST},
      url={https://arxiv.org/abs/2205.04341}, 
}

@article{singh2000convergence,
  title={Convergence results for single-step on-policy reinforcement-learning algorithms},
  author={Singh, Satinder and Jaakkola, Tommi and Littman, Michael L and Szepesv{\'a}ri, Csaba},
  journal={Machine learning},
  volume={38},
  pages={287--308},
  year={2000},
  publisher={Springer}
}

@misc{liu2025understandingr1zeroliketrainingcritical,
      title={Understanding R1-Zero-Like Training: A Critical Perspective}, 
      author={Zichen Liu and Changyu Chen and Wenjun Li and Penghui Qi and Tianyu Pang and Chao Du and Wee Sun Lee and Min Lin},
      year={2025},
      eprint={2503.20783},
      archivePrefix={arXiv},
      primaryClass={cs.LG},
      url={https://arxiv.org/abs/2503.20783}, 
}

@article{schulman2015high,
  title={High-dimensional continuous control using generalized advantage estimation},
  author={Schulman, John and Moritz, Philipp and Levine, Sergey and Jordan, Michael and Abbeel, Pieter},
  journal={arXiv preprint arXiv:1506.02438},
  year={2015}
}

@article{hendrycksmath2021,
    title={Measuring Mathematical Problem Solving With the MATH Dataset},
    author={Dan Hendrycks
    and Collin Burns
    and Saurav Kadavath
    and Akul Arora
    and Steven Basart
    and Eric Tang
    and Dawn Song
    and Jacob Steinhardt},
    journal={arXiv preprint arXiv:2103.03874},
    year={2021}
}

@article{li2024numinamath,
  title={Numinamath: The largest public dataset in ai4maths with 860k pairs of competition math problems and solutions},
  author={Li, Jia and Beeching, Edward and Tunstall, Lewis and Lipkin, Ben and Soletskyi, Roman and Huang, Shengyi and Rasul, Kashif and Yu, Longhui and Jiang, Albert Q and Shen, Ziju and others},
  journal={Hugging Face repository},
  volume={13},
  pages={9},
  year={2024}
}

@article{hendrycks2021measuring,
  title={Measuring mathematical problem solving with the math dataset},
  author={Hendrycks, Dan and Burns, Collin and Kadavath, Saurav and Arora, Akul and Basart, Steven and Tang, Eric and Song, Dawn and Steinhardt, Jacob},
  journal={arXiv preprint arXiv:2103.03874},
  year={2021}
}

@article{lewkowycz2022solving,
  title={Solving quantitative reasoning problems with language models},
  author={Lewkowycz, Aitor and Andreassen, Anders and Dohan, David and Dyer, Ethan and Michalewski, Henryk and Ramasesh, Vinay and Slone, Ambrose and Anil, Cem and Schlag, Imanol and Gutman-Solo, Theo and others},
  journal={Advances in Neural Information Processing Systems},
  volume={35},
  pages={3843--3857},
  year={2022}
}

@article{he2024olympiadbench,
  title={Olympiadbench: A challenging benchmark for promoting agi with olympiad-level bilingual multimodal scientific problems},
  author={He, Chaoqun and Luo, Renjie and Bai, Yuzhuo and Hu, Shengding and Thai, Zhen Leng and Shen, Junhao and Hu, Jinyi and Han, Xu and Huang, Yujie and Zhang, Yuxiang and others},
  journal={arXiv preprint arXiv:2402.14008},
  year={2024}
}

@article{abdolmaleki2024learning,
  title={Learning from negative feedback, or positive feedback or both},
  author={Abdolmaleki, Abbas and Piot, Bilal and Shahriari, Bobak and Springenberg, Jost Tobias and Hertweck, Tim and Joshi, Rishabh and Oh, Junhyuk and Bloesch, Michael and Lampe, Thomas and Heess, Nicolas and others},
  journal={arXiv preprint arXiv:2410.04166},
  year={2024}
}

@article{hu2025reinforce++,
  title={Reinforce++: A simple and efficient approach for aligning large language models},
  author={Hu, Jian},
  journal={arXiv preprint arXiv:2501.03262},
  year={2025}
}

@article{zhang2025rspo,
  title={RSPO: Risk-Seeking Policy Optimization for Pass@ k and Max@ k Metrics in Large Language Models},
  author={Zhang, Kaichen and Gao, Shenghao and Hong, Yuzhong and Sun, Haipeng and Bao, Junwei and Jiang, Hongfei and Song, Yang and Dingqian, Hong and Xiong, Hui},
  journal={arXiv preprint arXiv:2508.01174},
  year={2025}
}

@inproceedings{fedus2020revisiting,
  title={Revisiting fundamentals of experience replay},
  author={Fedus, William and Ramachandran, Prajit and Agarwal, Rishabh and Bengio, Yoshua and Larochelle, Hugo and Rowland, Mark and Dabney, Will},
  booktitle={International conference on machine learning},
  pages={3061--3071},
  year={2020},
  organization={PMLR}
}

@article{tokdar2010importance,
  title={Importance sampling: a review},
  author={Tokdar, Surya T and Kass, Robert E},
  journal={Wiley Interdisciplinary Reviews: Computational Statistics},
  volume={2},
  number={1},
  pages={54--60},
  year={2010},
  publisher={Wiley Online Library}
}

@article{qian2025beyond,
  title={Beyond the Next Token: Towards Prompt-Robust Zero-Shot Classification via Efficient Multi-Token Prediction},
  author={Qian, Junlang and Zhu, Zixiao and Zhou, Hanzhang and Feng, Zijian and Zhai, Zepeng and Mao, Kezhi},
  journal={arXiv preprint arXiv:2504.03159},
  year={2025}
}

@book{miao2024optimizing,
  title={Optimizing the Unknown: Reinforcement Learning and Energy-Based Model for Black Box Bayesian Optimization},
  author={Miao, Ruiyao},
  year={2024},
  publisher={University of California, Los Angeles}
}

@article{guo2025logic,
  title={Logic-in-frames: Dynamic keyframe search via visual semantic-logical verification for long video understanding},
  author={Guo, Weiyu and Chen, Ziyang and Wang, Shaoguang and He, Jianxiang and Xu, Yijie and Ye, Jinhui and Sun, Ying and Xiong, Hui},
  journal={arXiv preprint arXiv:2503.13139},
  year={2025}
}

@article{xu2025mark,
  title={Mark your llm: Detecting the misuse of open-source large language models via watermarking},
  author={Xu, Yijie and Liu, Aiwei and Hu, Xuming and Wen, Lijie and Xiong, Hui},
  journal={arXiv preprint arXiv:2503.04636},
  year={2025}
}

@inproceedings{jiang2024killing,
  title={Killing Two Birds with One Stone: Cross-modal Reinforced Prompting for Graph and Language Tasks},
  author={Jiang, Wenyuan and Wu, Wenwei and Zhang, Le and Yuan, Zixuan and Xiang, Jian and Zhou, Jingbo and Xiong, Hui},
  booktitle={Proceedings of the 30th ACM SIGKDD Conference on Knowledge Discovery and Data Mining},
  pages={1301--1312},
  year={2024}
}

@article{hu2024let,
  title={Let's Ask GNN: Empowering Large Language Model for Graph In-Context Learning},
  author={Hu, Zhengyu and Li, Yichuan and Chen, Zhengyu and Wang, Jingang and Liu, Han and Lee, Kyumin and Ding, Kaize},
  journal={arXiv preprint arXiv:2410.07074},
  year={2024}
}

@article{hu2024language,
  title={Language model preference evaluation with multiple weak evaluators},
  author={Hu, Zhengyu and Zhang, Jieyu and Xiong, Zhihan and Ratner, Alexander and Xiong, Hui and Krishna, Ranjay},
  journal={arXiv preprint arXiv:2410.12869},
  year={2024}
}

@article{hu2024explaining,
  title={Explaining length bias in llm-based preference evaluations},
  author={Hu, Zhengyu and Song, Linxin and Zhang, Jieyu and Xiao, Zheyuan and Wang, Tianfu and Chen, Zhengyu and Yuan, Nicholas Jing and Lian, Jianxun and Ding, Kaize and Xiong, Hui},
  journal={arXiv preprint arXiv:2407.01085},
  year={2024}
}

@article{hu2025unveiling,
  title={Unveiling the Learning Mind of Language Models: A Cognitive Framework and Empirical Study},
  author={Hu, Zhengyu and Lian, Jianxun and Xiao, Zheyuan and Zhang, Seraphina and Wang, Tianfu and Yuan, Nicholas Jing and Xie, Xing and Xiong, Hui},
  journal={arXiv preprint arXiv:2506.13464},
  year={2025}
}

@inproceedings{wang2025llm,
  title={Llm-powered multi-agent framework for goal-oriented learning in intelligent tutoring system},
  author={Wang, Tianfu and Zhan, Yi and Lian, Jianxun and Hu, Zhengyu and Yuan, Nicholas Jing and Zhang, Qi and Xie, Xing and Xiong, Hui},
  booktitle={Companion Proceedings of the ACM on Web Conference 2025},
  pages={510--519},
  year={2025}
}

@article{li2023remax,
  title={Remax: A simple, effective, and efficient reinforcement learning method for aligning large language models},
  author={Li, Ziniu and Xu, Tian and Zhang, Yushun and Lin, Zhihang and Yu, Yang and Sun, Ruoyu and Luo, Zhi-Quan},
  journal={arXiv preprint arXiv:2310.10505},
  year={2023}
}

@article{yue2025does,
  title={Does reinforcement learning really incentivize reasoning capacity in llms beyond the base model?},
  author={Yue, Yang and Chen, Zhiqi and Lu, Rui and Zhao, Andrew and Wang, Zhaokai and Song, Shiji and Huang, Gao},
  journal={arXiv preprint arXiv:2504.13837},
  year={2025}
}

@article{wu2025reasoning,
  title={Reasoning or Memorization? Unreliable Results of Reinforcement Learning Due to Data Contamination},
  author={Wu, Mingqi and Zhang, Zhihao and Dong, Qiaole and Xi, Zhiheng and Zhao, Jun and Jin, Senjie and Fan, Xiaoran and Zhou, Yuhao and Fu, Yanwei and Liu, Qin and others},
  journal={arXiv preprint arXiv:2507.10532},
  year={2025}
}

@article{shao2025spurious,
  title={Spurious rewards: Rethinking training signals in rlvr},
  author={Shao, Rulin and Li, Shuyue Stella and Xin, Rui and Geng, Scott and Wang, Yiping and Oh, Sewoong and Du, Simon Shaolei and Lambert, Nathan and Min, Sewon and Krishna, Ranjay and others},
  journal={arXiv preprint arXiv:2506.10947},
  year={2025}
}

@inproceedings{fan2025preference,
  title={Preference-Oriented Supervised Fine-Tuning: Favoring Target Model over Aligned Large Language Models},
  author={Fan, Yuchen and Hong, Yuzhong and Wang, Qiushi and Bao, Junwei and Jiang, Hongfei and Song, Yang},
  booktitle={Proceedings of the AAAI Conference on Artificial Intelligence},
  volume={39},
  number={22},
  pages={23859--23867},
  year={2025}
}

@article{stiennon2020learning,
  title={Learning to summarize with human feedback},
  author={Stiennon, Nisan and Ouyang, Long and Wu, Jeffrey and Ziegler, Daniel and Lowe, Ryan and Voss, Chelsea and Radford, Alec and Amodei, Dario and Christiano, Paul F},
  journal={Advances in neural information processing systems},
  volume={33},
  pages={3008--3021},
  year={2020}
}

@article{dai2025seper,
  title={Seper: Measure retrieval utility through the lens of semantic perplexity reduction},
  author={Dai, Lu and Xu, Yijie and Ye, Jinhui and Liu, Hao and Xiong, Hui},
  journal={arXiv preprint arXiv:2503.01478},
  year={2025}
}

@article{zhang2023impact,
  title={The impact of generative artificial intelligence on market equilibrium: Evidence from a natural experiment},
  author={Zhang, Kaichen and Yuan, Zixuan and Xiong, Hui},
  journal={arXiv preprint arXiv:2311.07071},
  year={2023}
}

@inproceedings{zhang2024optimized,
  title={Optimized Cost Per Click in Online Advertising: A Theoretical Analysis},
  author={Zhang, Kaichen and Yuan, Zixuan and Xiong, Hui},
  booktitle={Proceedings of the 30th ACM SIGKDD Conference on Knowledge Discovery and Data Mining},
  pages={4232--4243},
  year={2024}
}
